\documentclass[pdflatex,sn-mathphys]{sn-jnl}
\usepackage{comment}

\newcommand{\activities}{\mathcal{A}}
\newcommand{\multiset}{\mathcal{B}}

\jyear{2022}%

\theoremstyle{thmstyleone}%
\theoremstyle{thmstyletwo}%

\theoremstyle{thmstylethree}%
\newtheorem{definition}{Definition}%

\begin{document}

\title[Performance-Preserving Event Log Sampling for Predictive Monitoring]{Performance-Preserving Event Log Sampling for Predictive Monitoring\footnote{This article is a postprint and has been accepted for publication in the Journal of Intelligent Information Systems (to appear). DOI pending. This version is distributed according to the Creative Commons BY-NC license (\url{https://creativecommons.org/licenses/by-nc/4.0/}) and is \textcopyright the authors. The published version is \textcopyright Springer.}}


\author*[1]{\fnm{Mohammadreza} \sur{Fani Sani}}\email{fanisani@pads.rwth-aachen.de}

\author[2]{\fnm{Mozhgan} \sur{Vazifehdoostirani}}\email{m.vazifehdoostirani@tue.nl}

\author[1]{\fnm{Gyunam} \sur{Park}}\email{gnpark@pads.rwth-aachen.de}

\author[1]{\fnm{Marco} \sur{Pegoraro}}\email{pegoraro@pads.rwth-aachen.de}

\author[1,3]{\fnm{Sebastiaan J.} \sur{van Zelst}}\email{s.j.v.zelst@pads.rwth-aachen.de}

\author[1,3]{\fnm{Wil M.P.} \sur{van der Aalst}}\email{wvdaalst@pads.rwth-aachen.de}

\affil*[1]{\orgdiv{Process and Data Science Chair}, \orgname{RWTH Aachen University}, \orgaddress{\city{Aachen},\country{Germany}}}

\affil[2]{\orgdiv{Industrial Engineering and Innovation Science}, \orgname{Eindhoven University of Technology, Eindhoven}, \orgaddress{ \country{the Netherlands}}}

\affil[3]{\orgdiv{Fraunhofer FIT}, \orgaddress{ \city{Birlinghoven Castle, Sankt Augustin}, \country{Germany}}}


\abstract{Predictive process monitoring is a subfield of process mining that aims to estimate case or event features for running process instances. Such predictions are of significant interest to the process stakeholders. However, most of the state-of-the-art methods for predictive monitoring require the training of complex machine learning models, which is often inefficient.
Moreover, most of these methods require a hyper-parameter optimization that requires several repetitions of the training process which is not feasible in many real-life applications. In this paper, we propose an instance selection procedure that allows sampling training process instances for prediction models. We show that our instance selection procedure allows for a significant increase of training speed for next activity and remaining time prediction methods while maintaining reliable levels of prediction accuracy.}




\maketitle

\section{Introduction}\label{introduction}
The main goal of predictive process monitoring is to provide timely information by predicting the behavior of business processes~\cite{van_der_aalst_time_2011} and enabling proactive actions to improve the performance of the process~\cite{park_action-oriented_2022}.
It provides various predictive information such as the next performing activity of a process instance~\cite{hitfox_group_comprehensible_2016}, e.g., patient and product, its waiting time for an activity, its remaining time to complete the process, etc~\cite{marquez-chamorro_predictive_2018}.
For instance, by predicting the long waiting time of a patient for registration, one can bypass the activity or add more resources to perform it.

A plethora of approaches have been proposed to support predictive process monitoring. 
In particular, with the recent breakthroughs in machine learning, various machine learning-based approaches have been developed~\cite{marquez-chamorro_predictive_2018}.
The emergence of ensemble learning methods leads to improvement in accuracy in different areas~\cite{breiman1996bagging}.
eXtreme Gradient Boosting (XGBoost)~\cite{XGboost} has shown promising results, often outperforming other ensemble methods such as Random Forest or using a single regression tree ~\cite{senderovich2017intra,teinemaa2019outcome}.
Furthermore, techniques based on deep neural networks, e.g., Long-Short Term Memory (LSTM) networks, have shown high performance in 
different predictive tasks~\cite{DBLP:journals/dss/EvermannRF17}.



However, machine learning-based techniques are computationally expensive due to their training process~\cite{ZHOU2017350}.
Moreover, they often require exhaustive hyperparameter-tuning to provide acceptable accuracy.
Such limitations hinder the application of machine learning-based techniques into real-life business processes where new prediction models are required in short intervals to adapt to changing business situations.
Business analysts need to test the efficiency and reliability of their conclusions via repeated training of different prediction models with different parameters~\cite{marquez-chamorro_predictive_2018}. 
Such long training time limits the application of the techniques when considering the limitations in time and hardware~\cite{DBLP:conf/spaa/PourghassemiZLC20}. 


In this regard, \emph{instance selection} has been studied as a promising direction of research to reduce original datasets to a manageable volume to perform machine learning tasks, while the quality of the results (e.g., accuracy) is maintained as if the original dataset was used~\cite{10.5555/2671164}.
Instance selection techniques are categorized into two classes based on the way they select instances. 
First, some techniques select the instances at the boundaries of classes. 
For instance, Decremental Reduction Optimization Procedure (DROP)~\cite{10.1023/A:1007626913721} selects instances using \textit{k}-Nearest Neighbors by incrementally discarding an instance if its neighbors are correctly classified without the instance.
The other techniques preserve the instances residing inside classes, e.g., Edited Nearest Neighbor (ENN)~\cite{mnn} preserves instances by repeatedly discarding an instance if it does not belong to the class of the majority of its neighbors.


However, it is restricted to directly apply existing techniques for instance selection to predictive process monitoring training, since such techniques assume independence among instances~\cite{10.1023/A:1007626913721}.
Instead, in predictive process monitoring, instances are computed from event data that are recorded by the information system supporting business processes~\cite{de_leoni_general_2016}.
Thus, they are highly correlated~\cite{process-mining} with the notion of \textit{case}, e.g., a manufacturing product in a factory and a customer in a store.

In this work, we suggest an instance selection approach for predicting the next activity, the remaining time and the outcome of the process that are main applications of predictive business process monitoring. 
By considering the characteristics of the event data, the proposed approach samples event data such that the training speed is improved while the accuracy of the resulting prediction model is maintained.
We have evaluated the proposed methods using three real-life datasets and state-of-the-art techniques for predictive business process monitoring, including LSTM~\cite{LSTM} and XGBoost~\cite{XGboost}.

This paper extends our earlier work presented in~\cite{sani2021event} in the following dimensions: 1) Evaluating the applicability of the proposed approach, 2) Enhancing the accessibility of the work, and 3) Extending the discussion of strengths and limitations.
First, we have evaluated the applicability of the proposed approach both task-wise and domain-wise.
For the task-wise evaluation, we have selected the three most well-known predictive monitoring tasks (i.e., \emph{next activity}, \emph{remaining time}, and \emph{outcome} predictions) and evaluated the performance of the proposed approach in the different tasks.
For the domain-wise evaluation, we have evaluated the performance of our proposed approach in real-life event logs from different domains, including \emph{finance}, \emph{government}, and \emph{healthcare} domains.
Second, we have extended the accessibility of the proposed approach by implementing the proposed sampling methods in the Python platform as well as the Java platform.
Finally, we have extensively discussed the strengths and limitations of the proposed approach, providing foundations for further research.

The remainder is organized as follows. We discuss the related work in \autoref{sec:relatedwork}. Next, we present the preliminaries in \autoref{sec:preliminaries} and proposed methods in \autoref{sec:methods}. Afterward, \autoref{sec:eval} evaluates the proposed methods using real-life event data and \autoref{sec:disc} provides discussions. Finally, \autoref{sec:conc} concludes the paper.

\section{Related Work}\label{sec:relatedwork}
This section presents the related work on predictive process monitoring, time optimization, and instance sampling.
\subsection{Predictive Process Monitoring}

Predictive process monitoring is an exceedingly active field, both currently and historically, thanks to the compatibility of process sciences and other branches of data science that include inference techniques, such as statistics and machine learning. 
At its core, the fundamental component of many predictive monitoring approaches is the abstraction technique it uses to obtain a fixed-length representation of the process component subject to the prediction (often, but not always, process traces). In the earlier approaches, the need for such abstraction was overcome through model-aware techniques, employing process models and replay techniques on partial traces to abstract a flat representation of event sequences. Such process models are mostly automatically discovered from a set of available complete traces, and require perfect fitness on training instances (and, seldomly, also on unseen test instances). For instance, Van der Aalst et al.~\cite{van_der_aalst_time_2011} proposed a time prediction framework based on replaying partial traces on a transition system, effectively clustering training instances by control-flow information. This framework has later been the basis for a prediction method by Polato et al.~\cite{DBLP:journals/computing/PolatoSBL18}, where the transition system is annotated with an ensemble of SVR and Na{\"i}ve Bayes classifiers, to perform a more accurate time estimation. Some more recent approaches split the predictive contribution of process models and machine learning models in a perspective-wise manner: for instance, Park et al.~\cite{DBLP:journals/dss/ParkS20} obtain a representation of the performance perspective using an annotated transition system, and design an ensemble with a deep neural network to obtain the final predictive model. A related approach, albeit more linked to the simulation domain and based on a Monte Carlo method, is the one proposed by Rogge-Solti and Weske~\cite{DBLP:conf/icsoc/Rogge-SoltiW13}, which maps partial process instances in an enriched Petri net.

Recently, predictive process monitoring started to use a plethora of machine learning approaches, achieving varying degrees of success. For instance, Teinemaa et al.~\cite{teinemaa2016predictive} provided a framework to combine text mining methods with Random Forest and Logistic Regression. Senderovich et al.~\cite{senderovich2017intra} studied the effect of using intra-case and inter-case features in predictive process monitoring and showed a promising result for XGBoost compared to other ensemble and linear methods. A comprehensive benchmark on using classical machine learning approaches for outcome-oriented predictive process monitoring tasks~\cite{teinemaa2019outcome} has shown that the XGBoost is the best-performing classifier among different machine learning approaches such as SVM, Decision Tree, Random Forest, and logistic regression. 

More recent methods are model-unaware and perform based on a single and more complex machine learning model instead of an ensemble. In fact, such an evolution of predictive monitoring mimics the advancement in the accuracy of newer machine learning approaches, specifically the numerous and sophisticated models based on deep neural networks that have been developed in the last decade.
The LSTM network model has proven to be particularly effective for predictive monitoring~\cite{DBLP:journals/dss/EvermannRF17,DBLP:conf/caise/TaxVRD17}, since the recurrent architecture can natively support sequences of data of arbitrary length. It allows performing trace prediction while employing a fixed-length event abstraction, which can be based on control-flow alone~\cite{DBLP:journals/dss/EvermannRF17,DBLP:conf/caise/TaxVRD17}, data-aware~\cite{DBLP:conf/ssci/NavarinVPS17}, time-aware~\cite{DBLP:conf/icpm/NguyenCWSME20}, text-aware~\cite{DBLP:conf/bis/PegoraroUGA21}, or model-aware~\cite{DBLP:journals/dss/ParkS20}. Additionally, rather than leveraging control-flow information for prediction, some recent research aims to use predictive monitoring to reconstruct missing control-flow attributes such as labels~\cite{DBLP:journals/is/AaRL21} or case identifiers~\cite{DBLP:conf/bpmds/0001UHA22,DBLP:journals/corr/abs-2212-00009}. However, the body of work currently standing in predictive process monitoring, regardless of architecture of the predictive model and/or target of the prediction, does not include strategies for performance-preserving sampling.

\subsection{Time Optimization and Instance Sampling}

The latest research developments in the field of predictive monitoring, much like both this paper and the application of machine learning techniques in other domains, shifts its focus away from increasing the quality of prediction (with respect to a given error metric or a given benchmark), and specializes in enriching the results of the prediction on additional aspects or perspectives. Unlike the methods mentioned in the previous paragraph, this is usually obtained through dedicated machine learning models---or modifications thereof---rather than designing specific event- or trace-level abstractions. For instance, many scholars have attempted to make predictions more transparent through the use of explainable machine learning techniques~\cite{DBLP:conf/bpm/Verenich19,DBLP:conf/ecis/StierleBWZM021,DBLP:conf/bpm/SindhgattaMOB20,DBLP:conf/icpm/GalantiCLCN20,DBLP:journals/corr/abs-2210-16786}. More related to our present work, Pauwels and Calders~\cite{DBLP:conf/bpm/PauwelsC21} propose a technique to avoid the time expenditure caused by the retrain of machine learning models; this is necessary when they are not representative anymore---for instance, when changes occur in the underlying process (caused e.g. by concept drift). While in this paper we focus on the data, and we propose a solution based on sampling, Pauwels and Calders intervene on the model side, devising an incremental training schema which accounts for new information in an efficient way.

Another concept similar to the idea proposed in this paper, and of current interest in the field of machine learning, is \emph{dataset distillation}: utilizing a dataset to obtain a smaller set of training instances that contain the same information (with respect to training a machine learning model)~\cite{wang2020dataset}. While this is not considered sampling, since some instances of the distilled dataset are created ex-novo, it is an approach very similar to the one we illustrate in our paper.

The concept of instance sampling, or instance subset selection, is present in the context of process mining at large, albeit the development of such techniques is very recent. Some instance selection algorithms have been proposed to help classical process mining tasks. For example,~\cite{sani_2021_sampling} proposes to use instance selection techniques to improve the performance of process discovery algorithms; in this context, the goal is to obtain automatically a descriptive model of the process, on the basis of the data recorded about the historical executions of process instances---the same starting point of the present work. Then, the work in~\cite{sani_2020_editDistance} applies the same concept and integrates the edit distance to obtain a fast technique to approximate the conformance checking score of process traces: this consists of measuring the deviation between a model, which often represents the normative or prescribed behavior of a process, and the data, which represents the actual behavior of a process. This paper integrates the two aforementioned works, extending the effects of strategic sampling: while in~\cite{sani_2021_sampling} the sampling optimizes descriptive modeling and in~\cite{sani_2020_editDistance} it optimizes process diagnostics, in this work it aids predictive modeling.

To the best of our knowledge, no work present in literature inspects the effects of building a training set for predictive process monitoring through a strategic process instance selection, with the exception of our previous work, which we extend in this paper~\cite{sani2021event}. In this paper, we examine the underexplored topic of event data sampling and selection for predictive process monitoring, with the objective of assessing if and to which extent prediction quality can be retained when we utilize subsets of the training data.

\section{Preliminaries}\label{sec:preliminaries}
In this section, some process mining concepts such as event log and sampling are discussed.
In process mining, we use events to provide insights into the execution of business processes.
Event logs, i.e., collections of events representing the execution of several instances of a process, are the starting point of process mining algorithms.
An example event log is shown in \autoref{tab:SimpleEventLog}.
Each \textit{event} that relates to a row in the table is related to specific activities of the underlying process. 
Furthermore, we refer to a collection of events related to a specific process instance of the process as a \textit{case} (represented by the \emph{Case-id} column).
Both cases and events may have different attributes. 
An event log that is a collection of events and cases is defined as follows.

\begin{table}[tb]
\caption{Simple example of an event log. Rows capture \emph{events} recorded in the context of the execution of the process. 
An event describes at what point in time an activity was performed. Other data attributes may be available as well. 
}

\centering
\resizebox{0.925\textwidth}{!}{%
\begin{tabular}{ |c c c c c c| } 
 \hline
    Case-id & Event-id & Activity name & Starting time & Finishing time ...\\ 
 \hline
    $\vdots$ &$\vdots$ &$\vdots$ &$\vdots$ &$\vdots$ &$\dots$\\ 
    7 & 35 & Register(a) & 2021-01-02 12:23 & 2021-01-02 12:25 &$\dots$\\ 
    7 & 36 & Analyze Defect(b) & 2021-01-02 12:30 & 2021-01-02 12:40 &$\dots$\\ 
    7 & 37 & Inform User(g) & 2021-01-02 12:45 & 2021-01-02 12:47 &$\dots$\\ 
    8 & 39 & Register(a) & 2021-01-02 12:23 & 2021-01-02 13:15 &$\dots$\\ 
    7 & 40 & Test Repair(e) & 2021-01-02 13:05 & 2021-01-02 13:20 &$\dots$\\ 
    7 & 41 & Archive Repair(h) & 2021-01-02 13:21 & 2021-01-02 13:22 &$\dots$\\ 
    8 & 42 & Analyze Defect(b) & 2021-01-02 12:30 & 2021-01-02 13:30 &$\dots$\\ 
    $\vdots$ &$\vdots$ &$\vdots$ &$\vdots$ &$\vdots$& $\ddots$ \\ 
 \hline
\end{tabular}}
\label{tab:SimpleEventLog}
\end{table}
\begin{definition}[Event Log]
Let $\mathcal{E}$ be the universe of events, $\mathcal{C}$ be the universe of cases, $\mathcal{AT}$ be the universe of attributes, and $\mathcal{U}$ be the universe of attribute values.
Moreover, let $C{\subseteq}\mathcal{C}$ be a non-empty set of cases, let $E{\subseteq}\mathcal{E}$ be a non-empty set of events.
We define $(C,E,\pi_C, \pi_E )$ as an event log, where $\pi_C{:}C {\times}\mathcal{AT} {\pfun} \mathcal{U}  $ and $\pi_E{:}E {\times}\mathcal{AT} {\pfun} \mathcal{U} $.
Any event in the event log has a case, and thus, $\nexists_{e\in E} ( \pi_E(e, case) \not\in C)$ and $\bigcup\limits_{e\in E}(\pi_E(e, case)){=} C $.

Let $\mathcal{A}{\subseteq}\mathcal{U}$ be the universe of activities and $\mathcal{V}{\subseteq}\mathcal{A}^*$ be the universe of sequences of activities. 
For  any $e{\in} E$, function $\pi_E(e, activity){\in} \activities$, which means that any event in the event log has an activity. 
Moreover, for any $c{\in}C $ function $\pi_C(c, variant){\in} \mathcal{A}^*{\setminus} \{\langle \rangle\}$ that means any case in the event log has a variant. 
\end{definition}
Therefore, there are some mandatory attributes that are \textit{case} and \textit{activity} for events and \textit{variants} for cases. 
For example, for event with Event-id equals to $35$ in \autoref{tab:SimpleEventLog}, $\pi_E(e, case)){=} 7$ and  $\pi_E(e, activity){=}\text{Register(a)}$.

Variants are the sequence of activities that are presented in each case. For example, for case $7$ in \autoref{tab:SimpleEventLog}, the variant is $\langle a,b,g,e,h \rangle$ (for the simplicity we show each activity by a letter). 
Variant information plays an important role an in some process mining applications, e.g., process discovery and conformance checking, just this information is considered. 
In this regard, event logs are considered as a multiset of sequences of activities. 
In the following, a simple event log is defined. 

\begin{definition}[Simple event log]
Let $\mathcal{A}$ be the universe of activities and let the universe of multisets over a set $X$ be denoted by $\multiset(X)$.
A simple event log is $L{\in} \multiset(\activities^*)  $. 
Moreover, let $\mathcal{EL} $ be the universe of event logs and $EL{=}(C,E,\pi_C,\pi_E){\in} \mathcal{EL} $ be an event log.
We define function $sl{:}\mathcal{EL}{\to}\multiset(\activities^*)$
 returns the simple event log of an event log where $sl(EL){=}[\sigma^k \mid \sigma {\in}\{ \pi_C(c,variant) \mid c{\in}C  \} \wedge k{=}\Sigma_{c\in C} \big(\pi_C(c,variant){=} \sigma\big)   ] $.
The set of unique variants in the event log is denoted by $\overline{sl(EL)}{=}\{ \pi_C(c,variant) \mid c{\in}C  \}$.
\end{definition}
Therefore, $sl$ returns the multiset of variants in the event logs. 
Note that the size of a simple event log equals the number of cases in the event logs, i.e., $ {\mid}sl(EL){\mid}{=}{\mid}C{\mid}$

In this paper, we use sampling techniques to reduce the size of event logs. 
An event log sampling method is defined as follows. 

\begin{definition}[Event log sampling]
Let $\mathcal{EL}$ be the universe of event logs and $\activities$ be the universe of activities. 
Moreover, let $EL{=}(C,E,\pi_C,\pi_E){\in} \mathcal{EL} $ be an event log, we define function $\delta{:}\mathcal{EL}{\to} \mathcal{EL}  $ that returns the sampled event log where if $(C',E',\pi'_C, \pi'_E){=}\delta (EL)$, then $C'{\subseteq}C$, $E'{\subseteq}E$, $\pi'_E{\subseteq}\pi_E$, $\pi'_C{\subseteq}\pi_C$, and consequently, $\overline{sl(\delta(EL))} {\subseteq} \overline{sl(EL)}$. 
We define that $\delta$ is a variant-preserving sampling if $\overline{sl(\delta(EL))} {=} \overline{sl(EL)}$.
\end{definition}
In other words, a sampling method is variant-preserving if and only if all the variants of the original event log are presented in the sampled event log.

To use machine learning methods for prediction, we usually need to transfer each case to one or more features. 
The feature is defined as follows. 

\begin{definition} [Feature]
Let $\mathcal{AT}$ be the universe of attributes, $\mathcal{U}$ be the universe of attribute values, and $\mathcal{C}$ be the universe of cases. 
Moreover, let $AT{\subseteq}\mathcal{AT}$ be a set of attributes. 
A feature is a relation between a sequence of attributes' values for $AT$ and the target attribute value, i.e., $f{\in} (\mathcal{U}^{{\mid}AT{\mid}} {\times} \mathcal{U} ) $.
We define $\mathit{fe}{:}\mathcal{C}{\times}\mathcal{EL}{\to}\multiset(\mathcal{U}^{{\mid}AT{\mid}} {\times} \mathcal{U} )$ is a function that receives a case and an event log, and returns a multiset of features. 
\end{definition}
For the next and final activity prediction, the target attribute value should be an activity. However, for the remaining time prediction, the target attribute value is a numerical value.  
Moreover, a case in the event log may have different features. 
For example, suppose that we only consider the activities.
For the case $\langle a,b,c,d \rangle$, we may have $(\langle a \rangle, b)$, $(\langle a,b \rangle, c)$, and $(\langle a,b,c \rangle, d)$ as features. 
Furthermore, $\sum\limits_{c\in C} fe(c,EL)$ are the corresponding features of event log $EL{=}(C,E,\pi_C,\pi_E)$ that could be given to different machine learning algorithms. 
For more details on how to extract features from event logs please refer to \cite{qafari2020feature}.

\section{Proposed Sampling Methods}\label{sec:methods}
\begin{figure}[tb]
    \centering
    \includegraphics[width=1.01\textwidth]{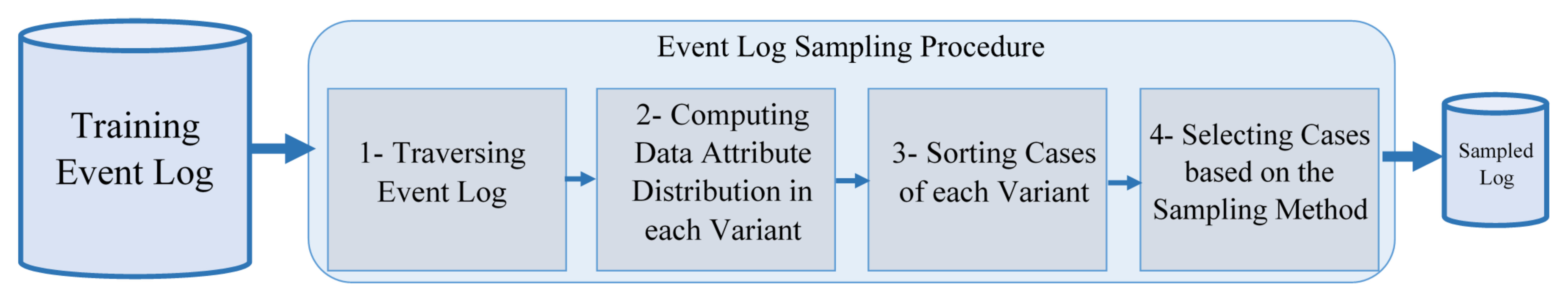}
    \caption{A schematic view of the proposed sampling procedure}
    \label{fig:samplingProcedure}
\end{figure}

In this section, we propose an event log preprocessing procedure that helps prediction algorithms to perform faster while maintaining reasonable accuracy. 
The schematic view of the proposed instance selection approach is presented in \autoref{fig:samplingProcedure}.
First, we need to traverse the event log and find the variants and corresponding traces of each variant in the event log. 
Moreover, different distributions of data attributes in each variant will be computed.  
Afterward, using different sorting and instance selection strategies, we are able to select some of the cases and return the sample event log. 
In the following, each of these steps is explained in more detail. To illustrate the following steps, we provide an example event log with 10 cases, visible in \autoref{tab:exampleLog}.

\begin{enumerate}
    \item \textit{Traversing the event log}: In this step, the unique variants of the event log and the corresponding traces of each variant are determined.
    In other words, consider event log $EL$ that $\overline{sl(EL)}{=}\{\sigma_1, \dots,\sigma_n \}$ where $n{=}{\mid} \overline{sl(EL)}{\mid}$, we aim to split $EL$ to $EL_1, \dots, EL_{n}$ where $EL_i$ only contains all the cases that $C_i{=}\{c{\in}C \mid \pi_C(c,variant){=} \sigma_i  \}$ and $E_i{=}\{e{\in}E \mid \pi_E(e,case){\in}C_i \}$. 
    Obviously, $\bigcup\limits_{1\leq i\leq n }(C_i){=}C$ and $\bigcap\limits_{1\leq i\leq n }(C_i){=}\varnothing$. 
    For the event log that is presented in \autoref{tab:exampleLog}, we have $n=4$ variants and $E_1=\{c_1,c_3,c_4,c_9, c_{10} \}$, $E_2=\{c_2,c_5, c_{8} \}$, $E_3=\{c6 \}$, and $E_1=\{c_7 \}$.  
    \item \textit{Computing Distribution}: In this step, for each variant of the event log, we compute the distribution of different data attributes $a{\in} AT$. 
    It would be more practical if the interesting attributes are chosen by an expert. Both event and case attributes can be considered. 
    A simple approach is to compute the frequency of categorical data values. 
    For numerical data attributes, it is possible to consider the average or the median of values for all cases of each variant. 
    In the running example for $E_3$ and $E_4$, we only have one case for each variant. However, for $E_1$, and $E_2$, the average of \textit{Amount} is $500$ and $460$, respectively.
    \item \textit{Sorting the cases of each variant}: In this step, we aim to sort the traces of each variant. 
    We need to sort the traces to give a higher priority to those traces that can represent the variant better. 
    One way is to sort the traces based on the frequency of the existence of the most occurred data values of the variant. 
    For example, we can give a higher priority to the traces that have more frequent resources of each variant.
    For the event log that is presented in \autoref{tab:exampleLog}, we do not need to prioritize the cases in $E_3$ and $E_4$. However, if we sort the traces according to their distance of amount value and the average value of each variant, for $E_1$, we have $c_3,c_9,c_4,c_1$, and $c_{10}$. The order for $E_2$ is $c_5,c_2$, and $c_8$.
    It is also possible to sort the traces based on their arrival time or randomly.    
    \item \textit{Returning sample event logs}: Finally, depending on the setting of the sampling function, we return some of the traces with the highest priority for all variants.
    The most important point about this step is to know how many traces of each variant should be selected. 
    
    \begin{table}[tb]
\caption{An example event log with 10 traces and 4 variants. Each trace has two attributes that are \textit{Variant} and \textit{Amount}.\label{tab:exampleLog}}
\centering
\begin{tabular}{|l|l|l|l}
\hline

\textbf{CaseID} & \textbf{Variant} & \textbf{Amount} \\ \hline
$c_1$ & $\langle a,b,c,d \rangle$ & 100 \\ \hline
$c_2$ &$\langle a,c,b,d \rangle$& 720 \\ \hline
$c_3$ &$\langle a,b,c,d \rangle$& 400 \\ \hline
$c_4$ &$\langle a,b,c,d \rangle$& 800 \\ \hline
$c_5$ &$\langle a,c,b,d \rangle$& 600 \\ \hline
$c_6$ &$\langle a,c,c,d \rangle$& 750 \\ \hline
$c_7$ &$\langle a,c,d \rangle$& 170 \\ \hline
$c_8$ &$\langle a,c,b,d \rangle$& 60 \\ \hline
$c_9$ &$\langle a,b,c,d \rangle$& 260 \\ \hline
$c_{10}$ &$\langle a,b,c,d \rangle$& 940 \\ \hline
\end{tabular}
\end{table}
    In the following, some possibilities will be introduced. 
    \begin{itemize}
        \item \textit{Unique selection}: In this approach, we select only one trace with the highest priority. In other words, suppose that $L'{=}sl(\delta(EL))$, $\forall_{\sigma\in L'} L'(\sigma){=}1 $.
        Therefore, using this approach we will have ${\mid}sl(\delta(EL)){\mid}{=} {\mid}\overline{sl(EL)}{\mid}$. It is expected that by using this approach, the distribution of frequency of variants will be changed and consequently the resulted prediction model will be less accurate.
        By applying this sampling method on the event log that is presented in \autoref{tab:exampleLog}, the sampled event log will have $4$ traces, i.e., one trace for each variant. The corresponding cases are $C'=\{ c_3, c_5,c_6,c_7\}$. For the variants that have more than one trace, the traces are chosen that have the highest priority (their amount value is closer to the average amount value for each variant).
        \item \textit{Logarithmic distribution}: In this approach, we reduce the number of traces in each variant in a logarithmic way. If $L{=}sl(EL)$ and $L'{=}sl(\delta(EL))$, $\forall_{\sigma\in L'} L'(\sigma){=}{[}Log_{k}(L(\sigma)) {]} $.
        Using this approach, the infrequent variants will not have any trace in the sampled event log and consequently it is not variant-preserving.
        According the above formula, by using a higher base for the logarithm (i.e., $k$), the size of the sampled event log is reduced more. 
        By using this sampling strategy with $k$ equals to $3$ on the event log that is presented in \autoref{tab:exampleLog}, the cases that selected in the sampled event log is $C'=\{ c_3, c_9,c_5\}$. Note that for the infrequent variants no trace is selected in the sampled event log.
        \item \textit{Division}: This approach performs similar to the previous one, however, instead of using logarithmic scale, we apply the division operator. In this approach, $\forall_{\sigma\in L'} L'(\sigma){=}{\lceil}\frac{L(\sigma)}{k} {\rceil} $.
        A higher $k$ results in fewer cases in the sample event log. 
        Note that as ${\lceil} {\rceil}$ considered in the above formula, using this approach all the variants have at least one trace in the sampled event log and it is variant-preserving.
        By using this sampling strategy with $k=4$ on the event log that is presented in \autoref{tab:exampleLog}, the sampled event log will have $5$ traces that are $C'=\{ c_3, c_9,c_5,c_6,c_7\}$.
    \end{itemize}
    There is also a possibility to consider other selection methods. For example, we can select the traces completely randomly from the original event log.
\end{enumerate}

By choosing different data attributes in Step 2 and different sorting algorithms in Step 3, we are able to lead the sampling of the method on \textit{which} cases should be chosen. 
Moreover, by choosing the type of distribution in Step 4, we determine \textit{how many} cases should be chosen.
To compute how sampling method $\delta$ reduces the size of the given event log $EL$, we use the following equation:
\begin{equation}
R_{S}{=}\frac{{\mid}sl(EL){\mid}}{{\mid}sl(\delta(EL)){\mid}}
\end{equation}
The higher $R_{S}$ value means, the sampling method reduces more the size of the training log. 
By choosing different distribution methods and different \textit{k-values}, we are able to control the size of the sampled event log.
It should be noted that the proposed method will apply just to the training event log. In other words, we do not sample event logs for development and test datasets.

\section{Evaluation}\label{sec:eval}
In this section, we aim at designing some experiments to answer the research question, i.e., ''Is it possible to have computational performance improvement of prediction methods by using the sampled event logs, while maintaining a similar accuracy?".
It should be noted that the focus of the experiments is not on prediction model tuning to have higher accuracy. Conversely, we aim to analyze the effect of using sampled event logs (instead of the whole datasets) on the required time and the accuracy of prediction models.  

In the following, we first explain the evaluation settings and event logs that are used. 
Afterward, we provide some information about the implementation of sampling methods, and finally, we show the experimental results.

\subsection{Evaluation Setting}
In this section, we first explain the prediction methods and parameters that are used in the evaluation. Afterward, we discuss the evaluation metrics. 

\subsubsection{Evaluation Parameters}

We have developed the sampling methods as a plug-in in the ProM framework~\cite{Prom}, accessible via \url{https://svn.win.tue.nl/repos/prom/Packages/LogFiltering}. 
This plug-in takes an event log and returns k different train and test event logs in the CSV format.
Moreover, we have also implemented the sampling methods in Python to have all the evaluations in one workflow. 

We have used two machine learning methods to train the prediction models, i.e., \textit{LSTM} and \textit{XGBoost}. 
For predicting the next activity, our LSTM network consisted of an input layer, two LSTM layers with \textit{dropout} rates of $10\%$, and a dense output layer with the \textit{SoftMax} activation function. We used “categorical cross-entropy” to calculate the loss and adopted \textit{ADAM} as an optimizer. We built the same architecture of \textit{LSTM} for the remaining time predicting with some differences. We employed “mean absolute error” as a loss function and “Root Mean Squared Propagation” \textit{(RMSprop)} as an optimizer. 
We used \textit{gbtree} with a max depth of $6$ as a booster in our XGBoost model for both of the next activity and remaining time prediction tasks. Uniform distribution is used as the sampling method inside our XGBoost model. To avoid overfitting in both models, the training set is further divided into $90\%$ training set and $10\%$ validation set to stop training once the model performance on the validation set stops improving. We used the same parameter setting of both models for original event logs and sampled event logs. 
The implementations of these methods are available at \url{https://github.com/gyunamister/pm-prediction/}.

To train the prediction models using machine learning methods, we extract features from event data.
To this end, we use the most commonly-used features for each prediction task in order to reduce the degree of freedom in selecting relevant features.
In other words, we focus on comparing the performance of predictions between sampled and non-sampled event data with a fixed feature space.
For instance, for the next activity prediction, we use the partial trace (i.e., the sequence of historical activities) of cases and the temporal measures of each activity (e.g., sojourn time) with one-hot encoding~\cite{DBLP:conf/icpm/ParkS19}.
For the remaining time prediction, we use the partial trace of cases along with case attributes (e.g., cost), resources, and temporal measures~\cite{verenich2019survey}.


To sample the event logs, we use three distributions that are \textit{$log$ distribution}, \textit{division}, and \textit{unique variants}. 
For the $log$ distribution method, we have used $2,3,5$, and $10$ (i.e., $log_2, log_3, log_5$, and $log_{10}$).
For the division method, we have used $2,3,5$, and $10$ (i.e., $d2,d3, d5$, and $d10$). 
For each event log and each sampling method, we have used a $5\text{-}fold$ cross-validation. 
It means we split the data into $5$ groups. One of the groups is used as the test event log, and the rest are merged as the training event log. It should be noted that for each event log, the splitting groups were the same for all the prediction and sampling methods.
Moreover, as the results of the experiments are non-deterministic, all the experiments have been repeated $5$ times, and the average values are represented.
Moreover, to have a fair evaluation, in all the steps, one CPU thread has been used.
\subsubsection{Metrics}
To evaluate the correctness of prediction methods for predicting the next activities, we have considered two metrics, i.e., \textit{Accuracy} and \textit{F1-score}. 
The F1-score is used for imbalanced data~\cite{DBLP:journals/pr/LuqueCMH19}. 
For remaining time prediction, we consider Mean Absolute Error (MAE) and Root Mean Squared Error (RMSE) measures as they were used in \cite{verenich2019survey} that are computed as follows. 
\begin{equation}
    MAE {=}\frac{1}{n}\sum_{t=1}^{n}{\mid}e_t{\mid}
\end{equation}
\begin{equation}
 RMSE {=}\sqrt{\frac{1}{n}\sum_{t=1}^{n}e_t^2}
\end{equation}
In the above equations, $e_t$ indicates the prediction error for the $t^{th}$ instance of validation data.
In other words, we aim to compute the absolute difference between the predicted and real values (in days) for the remained time. 
For both measures, the lower value means higher accuracy.  
Note that, similar to \cite{verenich2019survey}, we considered seconds as the time unit to compute these two metrics. 

To evaluate how accurate the prediction methods using the sampled event logs are, we have used relative metrics that compared them with the case that whole event logs are used according to the following equations.
\begin{equation}
    R_{Acc}=\frac{\text{ Accuracy using the sampled training log}}{\text{Accuracy using the whole training log}}
\end{equation}
\begin{equation}
     R_{F1}{=}\frac{\text{ F1-score using the sampled training log}}{\text{F1-score using the whole training log}}
\end{equation}
In both above equations, a value close to $1$ means that using the sampling event logs, the prediction methods behave almost similar to the case that the whole data is used for the training. Moreover, values higher than $1$ indicate the accuracy/F1-score of prediction methods has improved. 

Unlike previous metrics, for $MAE$ and $RMSE$, a higher value means the prediction model is less accurate in predicting the remaining time. 
Therefore, we use the following measures. 

\begin{equation}
    R_{MAE}{=}\frac{MAE \text{ using the whole training log}}{MAE \text{ using the sampled training log}}
\end{equation}
    
\begin{equation}
    R_{RMSE}{=}\frac{RMSE \text{ using the whole training log}}{RMSE \text{ using the sampled training log}}
\end{equation}

In both of the above measures, a higher value means that the applied instance selection method preserves higher accuracy. 
In case the values of the above measures are higher than $1$, the instance selection methods improve the accuracy of prediction models compared to the case that the whole training data have been used.

 To compute the improvement in the performance of feature extraction and training time, we will use the following equations. 
  
  \begin{equation}
        R_{FE}{=}\frac{\text{Feature extraction time using whole data}}{\text{Feature extraction time using the sampled data}}
    \end{equation}
    
    \begin{equation}
        R_{t}{=}\frac{\text{Training time using whole data}}{\text{Training time using the sampled data}}
    \end{equation}

    For both equations, the resulting values indicate how many times applying the sampled log is faster than using all data.

\subsubsection{Event Logs}

We have used three event logs widely used in the literature.
In the \textit{BPIC-2012-W} event log, relating to a process of an insurance company, the average of variant frequencies is low.
In the \textit{RTFM} event log, which corresponds to a road traffic management system, we have some highly frequent variants and several infrequent variants. Moreover, the number of activities in this event log is high.
In the \textit{Sepsis} event log, relating to a health care process, there are several variants, that most of them are unique. 
Some of the activities in the last two event logs are infrequent, which makes these event logs imbalanced.
Some information about these event logs and the result of using prediction methods on them is presented in \autoref{tab:eventLogs}. Note that the time-related features in this table are in seconds. 
\begin{table}[b]
     \centering
     \caption{Event logs that are used in the evaluation and results of using them for the next activity and remaining time prediction\label{tab:eventLogs}}
    \includegraphics[width=1.01\textwidth]{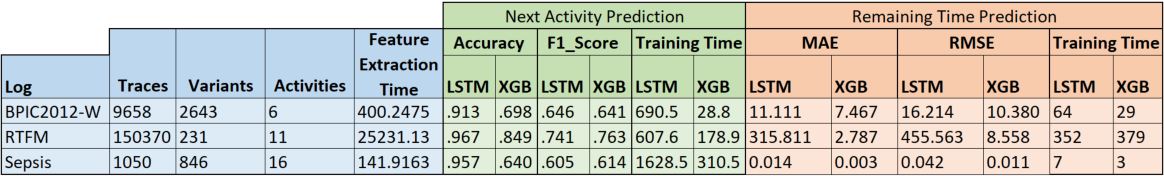}
\end{table}

According to \autoref{tab:eventLogs}, using the whole event data we usually have high accuracy for the next activity prediction. However, the F1-score is not that high, which is mainly because the event logs are imbalanced (specifically \textit{RTFM} and \textit{Sepsis}). 
Moreover, the MAE and RMSE values are very high. Specifically for the \textit{RTFM} event log. It is mainly because process instances' durations are very long in this event log, and consequently, the $e_t$ values are higher. 
Finally, there is a direct relation between the size of event logs and the required time for extracting features and training the models. 

\subsection{Evaluation Results}    
Here, we provide the results of using sampled training event logs instead of whole training event logs. 
First, we show how by using sampling, the size of training data is reduced in \autoref{tab:RS}. 
As it is expected, the highest reduction occurs when $log_{10}$ is used.
Using this sampling, the size of the \textit{RTFM} event log is more than $1000$ times reduced. 
However, for the $Sepsis$ event log, as most variants are unique, sampling the training event logs using divide distribution could not result in high $R_S$.
Moreover, this table shows how using the sampling event logs can reduce the required time to extract features of the event data, i.e., $R_{FE}$. 
As it is expected, there is a correlation between the size reduction of the sampled event logs and the improvement in the $R_{FE}$.
\begin{table}
     \centering
     \caption{Reduction in size of training event logs (i.e., $R_S$) and the improvement in the feature extraction process (i.e., $R_{FE}$) using different sampling methods\label{tab:RS}}
    \includegraphics[width=1.01\textwidth]{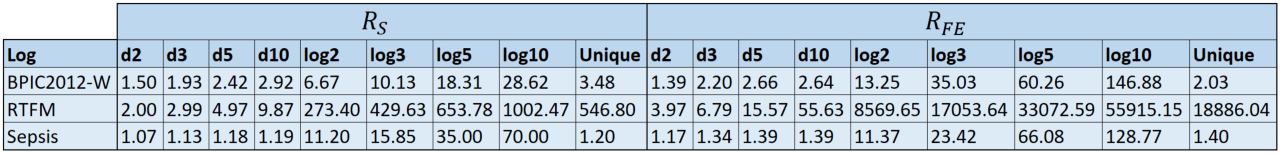}
\end{table}

In the following, we show how using the sampling event logs affects the next activity, remaining time, and outcome prediction. 

\subsubsection{Next Activity Prediction}
The accuracy of both LSTM and XGboost methods that are trained using sampled training data is presented in \autoref{tab:Acc}.
Results indicate that in most cases, when the division sampling methods are used, we can achieve similar accuracy, i.e., $R_{Acc}$ close to $1$, compared to the case where the whole training data is used. 
In some cases, like using $d2$ for the \textit{Sepsis} event log and LSTM method, the accuracy of the training method is even (slightly) improved. 
However, for the \textit{RTFM}, using sampling with logarithmic or unique distribution highly changes the frequency distribution of variants and consequently causes a higher reduction in the accuracy of predicting models.
Moreover, the accuracy reduction was higher when the XGboost method was used.
The result indicated that by increasing the $R_S$ value, we lose more information in the training event logs, and consequently, the quality of the prediction models will be decreased.

\begin{table}
     \centering
     \caption{$R_{Acc}$ of different event logs when different sampling methods are used.\label{tab:Acc}}
    \includegraphics[width=1.01\textwidth]{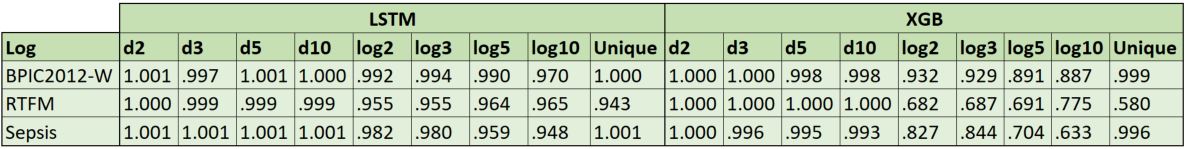}
\end{table}

In \autoref{tab:F1}, $R_{F1}$ of trained models are depicted. 
The results again indicate that using the sampling method with divide distribution in most cases leads to having a similar (and sometimes higher) F1-score. 

\begin{table}
     \centering
     \caption{$R_{F1}$ of different event logs when different sampling methods are used.\label{tab:F1}}
    \includegraphics[width=1.01\textwidth]{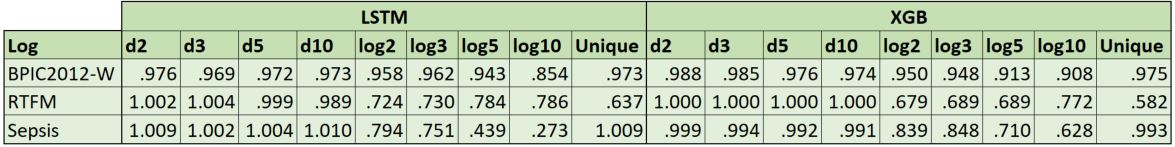}
\end{table} 
By considering the results of these two tables, we found that specifically, when the LSTM method is used, we will have similar accuracy and F1-score. 
Moreover, for the \textit{Sepsis} and \textit{BPIC2012-W} that variants have similar frequency having all the variants (i.e., using divide and unique distributions) can help the prediction method to have results similar to the case that we use the whole training data. 
However, for the \textit{RTFM} event log that has some high frequent variants, using the unique distribution results in lower accuracy and F1-score. 

\autoref{tab:TT} shows how much training time is faster using the sampled training data instead of using the whole event data. 
There is a direct relationship between the size reduction and $R_t$(refer to the results in \autoref{tab:RS}). However, in most cases, the performance improvement is bigger for the XGboost method. 
Considering the results in this table and \autoref{tab:F1}, we found that using the sampling method, we are able to improve the performance of the next activity prediction methods on the used event logs while they provide similar results. However, oversampling (e.g., applying $log10$ for \textit{RTFM}) will result in lower accuracy. 
\begin{table}
     \centering
     \caption{$R_{t}$ of different event logs when different sampling methods are used.\label{tab:TT}}
     
    \includegraphics[width=1.00\textwidth]{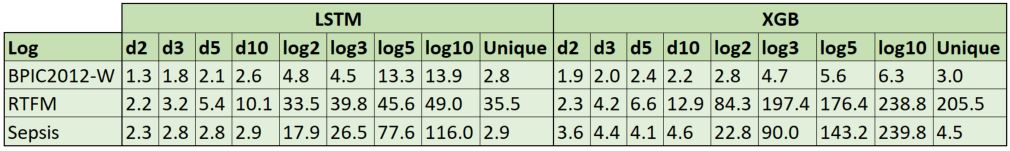}
\end{table}

\subsubsection{Remaining Time Prediction}
In \autoref{tab:MAE} and \autoref{tab:RMSE}, we show how by using the sampled event logs, MAE and RMSE of different remaining time prediction methods are changed. 
The results indicate that for the LSTM method, independent of the sampling method for all event logs, we are able to provide a prediction similar to the case where whole event logs are used. It seems that the settings that are used for training the prediction method are not good. In other words, the trained model is not accurate enough. We repeat this experiment for LSTM with several different parameters, but we have almost the same results. It is mainly caused by the challenging nature of the remaining time prediction task compared to classification-based problems (such as next activity and outcome prediction). However, by sampling the training event logs, we keep the quality of prediction models.   

For the XGboost method, the results indicate that if we do not sample a small amount of traces (for example, using logarithmic sampling), we can have high $R_{MAE}$ and $R_{RMSE}$. 
In general, as the main attribute for the next activity prediction is the sequence of activities, it is less sensitive to sampling. 
However, to predict the remaining time, the other data attributes can be essential too. 
In other words, for the remaining prediction, we need larger sampled event logs.

\begin{table}
     \centering
     \caption{$R_{MAE}$ of different event logs when different sampling methods are used.\label{tab:MAE}}
    \includegraphics[width=1.01\textwidth]{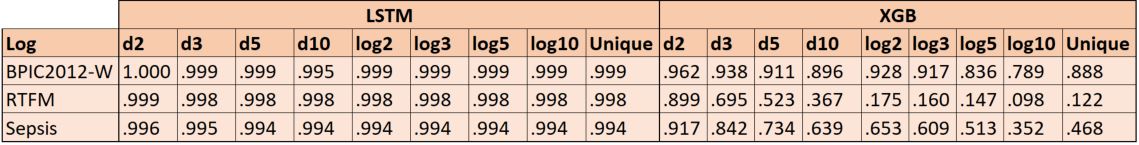}
\end{table}

\begin{table}
     \centering
     \caption{$R_{RMSE}$ of different event logs when different sampling methods are used.\label{tab:RMSE}}
     
    \includegraphics[width=1.01\textwidth]{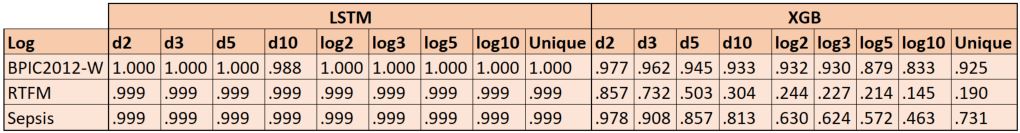}
\end{table}

In \autoref{tab:ReamainTT}, it is shown how by sampling event logs, we are able to reduce the required training time and improve the performance of the remaining time prediction process. 
By considering the results in \autoref{tab:ReamainTT} and \autoref{tab:TT}, as we have expected, by having higher $R_S$ the $R_t$ value is higher. 
\begin{table}
     \centering
     \caption{$R_{t}$ of different event logs when different sampling methods are used.\label{tab:ReamainTT}}
     
    \includegraphics[width=1.01\textwidth]{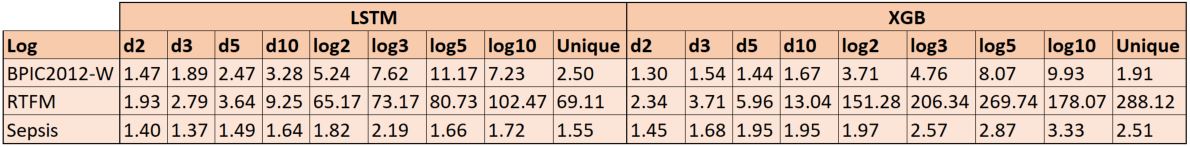}
\end{table}

\subsection{Outcome Prediction}
For the outcome prediction, in order to facilitate comparison and remain consistent with previous work on outcome prediction, we transform each event log into different event logs~\cite{Irene_outcome_2019}.
For example, we transform $BPIC_-2012$ event log to $BPIC_-2012_-Accepted$, $BPIC_-2012_-Cancelled$, and $BPIC_-2012_-Declined$.

The $R_{Acc}$ and $R_{F1}$ of both LSTM and XGboost methods that are trained for prediction of outcome using sampled training data is presented in \autoref{tab:finalAcc} and \autoref{tab:finalF1}, respectively.
The results indicate that, in many cases, we are able to improve the accuracy of the outcome prediction algorithms. Specifically, using \textit{Unique} strategy for sampling the traces for $BPIC_-2012_-Accepted$ event log leads to considerable improvement for both $LSTM$ and $XGB$ methods. 
Unlike the other two applications, i.e., the next activity and the remaining time prediction, even by oversampling some event logs, e.g., $log10$, we can obtain results similar to the cases in which the whole training event logs are used. 

In \autoref{tab:finalRT}, $R_t$ of different sampling methods are shown. The performance improvement is usually bigger for the \textit{LSTM} method. There are several cases in which we are not able to improve the performance of the prediction method ($R_t$ values less than $1$). It happens mainly for the \textit{Unique} sampling method. One reason could be by removing the frequencies, the convergence time for the learning method is increased. In case the logarithmic method is used, the performance improvement is around $50$ times. It means the training process using the sampled event log is $50$ times faster than the case where the whole training log is used. 

\begin{table}
     \centering
     \caption{$R_{Acc}$ of different event logs when different sampling methods are used for outcome prediction.\label{tab:finalAcc}}
    \includegraphics[width=1.01\textwidth]{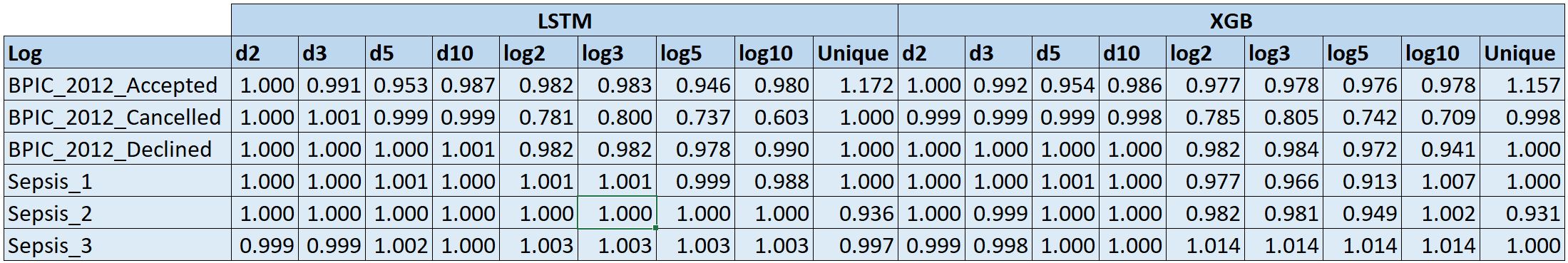}
\end{table}

\begin{table}
     \centering
     \caption{$R_{F1}$ of different event logs when different sampling methods are used for outcome prediction.\label{tab:finalF1}}
     
    \includegraphics[width=1.01\textwidth]{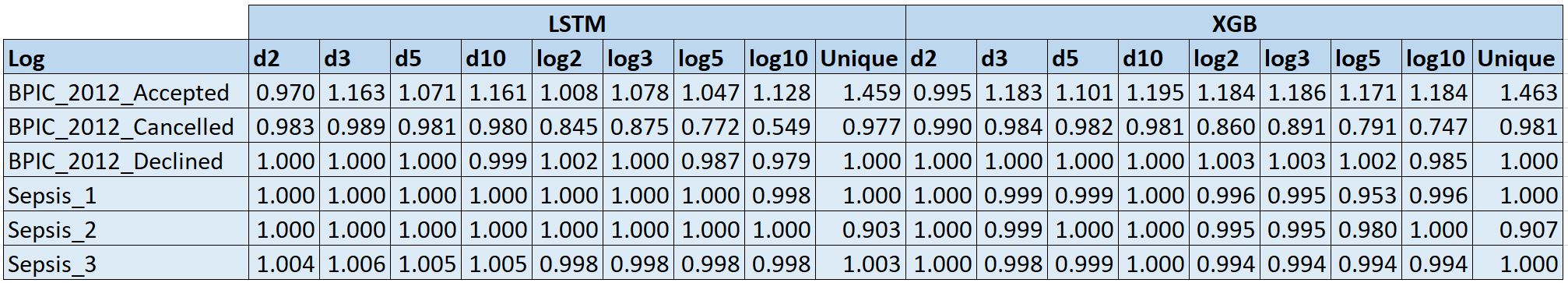}
\end{table}

\begin{table}
     \centering
     \caption{$R_{t}$ of different event logs when different sampling methods are used for outcome prediction.\label{tab:finalRT}}
     
    \includegraphics[width=1.01\textwidth]{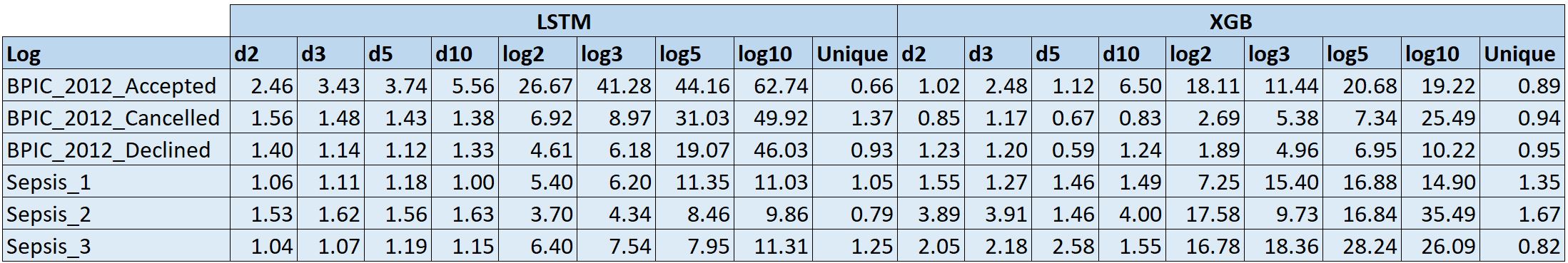}
\end{table}

\section{Discussion}\label{sec:disc}
In this section, we discuss the results that are illustrated in the previous section. The results indicate that we do not always have a typical trade-off between the accuracy of the trained model and the performance of the prediction procedure. For example, for the next activity prediction, there are some cases where the training process is much faster than the normal procedure, even though the trained model provides an almost similar or higher accuracy and F1-score. 
Thus, the proposed instance selection procedure can be applied when we aim to apply hyper-parameter optimization~\cite{Hyperparameter}. 
In this way, more settings can be analyzed in a limited time. Moreover, it is reasonable to use the proposed method when we aim to train an online prediction method or on naive hardware such as cell phones.

To achieve the highest performance improvement while the trained model is accurate enough, different sampling methods should be used for different event logs.  
For example, for the \textit{RTFM} event log---as there are some highly frequent variants---the division distribution may be more useful. In other words, independently of the used prediction method, if we change the distribution of variants (e.g., using $unique$ distribution), it is expected that the accuracy will sharply decrease.
However, for event logs with a more uniform distribution, we can use $unique$ distributions to sample event logs.
Furthermore, the results indicate that the effect of the chosen distribution (i.e., $unique$, $division$, and $logarithmic$) is more important than the used \textit{k-value}. 
It is mainly because the $logarithmic$ distribution may remove some of the variants, and the $unique$ distribution change the frequency distribution of variants. 
Therefore, it would be interesting to investigate more on the characteristics of the given event log and suitable sampling parameters for such distribution.
For example, if most variants of a given event log are unique (e.g., $Sepsis$), using the $logarithmic$ distribution leads to having remarkable $R_{S}$ and consequently, $R_{FE}$ and $R_{t}$ will be very high. However, we will lose most of the variants, and the trained model might have poor predictions.

By analyzing the results, we found that the infrequent activities can be ignored using some hyper-parameter settings. 
The significant difference between F1-score and Accuracy values in \autoref{tab:eventLogs} indicates this problem too. 
Using the sampling methods that modify the distribution of the event logs such as the $unique$ method can help the prediction methods to also consider these activities. 
However, as these activities are infrequent, improving in the prediction of them would not impact highly on the presented aggregated F1-score value.

Finally, in real-life business scenarios, the process can change because of different reasons~\cite{josep_2012_drift}. 
This phenomenon is usually called \textit{concept drift}. 
By considering the whole event log for training the prediction model, it is most probable that these changes are not considered in the prediction. Using the proposed sampling procedure, and giving higher priorities to newer traces, it is expected that we are able to adapt to the changes faster, which may be critical for specific applications.   

\subsection*{Limitations}
Comparing the results for the next activity and remaining time prediction, we found that predicting the remaining time of the process is more sensitive to instance selection. In other words, this application requires more data to predict accurately. 
Thus, for the cases that the target attribute depends more on other data attributes compared to variant information, we need to sample more data to capture more related information. Otherwise, the trained model might be inaccurate. 

We also found a critical problem in predictive monitoring. In some cases, specifically using LSTM for predicting the remaining time, the accuracy of the predictions is low. 
For the next activity prediction, it is possible that the prediction models almost ignore infrequent activities.
In these cases, even if we use the training data for the evaluation, we do not have acceptable results. This problem in machine learning is called a high bias error \cite{bias}. 
In other words, the training is not efficient even when using whole data and we need to change the prediction method (or its parameters).


\section{Conclusion}\label{sec:conc}
In this paper, we proposed an instance selection approach to improve the performance of predictive business process monitoring methods. We suggested that it is possible to use a sample of training event data instead of the whole training event data. 
To evaluate the proposed approach, we consider two main applications of predictive business monitoring, i.e., the next activity and the remaining time prediction. 
Results of applying the proposed approaches on three real-life event logs and two widely used machine learning methods, i.e., LSTM and XGboost, indicate that in most cases, we are able to improve the performance of predictive monitoring algorithms while providing similar accuracy compared to the case that the whole training event logs are used. 
However, by oversampling, the accuracy of the trained model might be reduced. 
Moreover, we have found that the remaining time prediction application is more sensitive to sampling. 

To continue this research, we aim to extend the experiments to study the relationship between the event log characteristics and the sampling parameters. 
In other words, we aim to help the end-user to adjust the sampling parameters based on the characteristics of the given event log. 
Moreover, it would be great to investigate how we can apply the proposed sampling procedure for streaming event data which is potentially one of the major advantages of the proposed method in a real-life setting. 
Finally, it is interesting to investigate more on feature selection methods for improving the performance of the predictive monitoring procedure. 
It is expected that, similar to process instance sampling, feature selection methods are able to reduce the required training time. 
In other words, the training is not efficient even using whole data and we need to change the prediction method (or its parameters).

Another important outcome of the results is that for different event logs, we should use different sampling methods to achieve the highest performance. Considering lots of different machine learning methods and their parameters can lead to an increase in the search space and complexity for users. 
In other words, finding the right setting for sampling may be challenging in real scenarios. 
Therefore, it would be valuable to research the relationship between the event log characteristics and suitable sampling parameters that can be used for preprocessing the training event log.

\section*{Acknowledgment}

The authors would like to thank the Alexander von Humboldt (AvH) Stiftung for funding this research.

\section*{Declarations}

In this part, we provide some declarations about the conflict of interest, the code availability, and the availability of data that is used in this paper. 

\begin{itemize}

\item Conflict of interest:

    \item Code availability: 
    Our proposed are available in \url{https://svn.win.tue.nl/repos/prom/Packages/LogFiltering} and \url{https://github.com/gyunamister/pm-prediction/}.
    For a part of the experiments, we have used the implementation that is available at \url{https://github.com/verenich/time-prediction-benchmark}.
    \item Data availability:
    We have applied our proposed approach to the following three publicly available datasets (event logs).
    \begin{itemize}
    	\item \textit{BPIC-2012}, that is accessible via \url{https://data.4tu.nl/articles/dataset/BPI_Challenge_2012/12689204}. 
    	\item \textit{RTFM}, that is accessible via \url{https://data.4tu.nl/articles/dataset/Road_Traffic_Fine_Management_Process/12683249}.
    	\item \textit{Sepsis} that is accessible via \url{https://data.4tu.nl/articles/dataset/Sepsis_Cases_-_Event_Log/12707639}.
    \end{itemize} 
\end{itemize}
\bibliography{sn-bibliography}


\begin{thebibliography}{47}
\ifx \bisbn   \undefined \def \bisbn  #1{ISBN #1}\fi
\ifx \binits  \undefined \def \binits#1{#1}\fi
\ifx \bauthor  \undefined \def \bauthor#1{#1}\fi
\ifx \batitle  \undefined \def \batitle#1{#1}\fi
\ifx \bjtitle  \undefined \def \bjtitle#1{#1}\fi
\ifx \bvolume  \undefined \def \bvolume#1{\textbf{#1}}\fi
\ifx \byear  \undefined \def \byear#1{#1}\fi
\ifx \bissue  \undefined \def \bissue#1{#1}\fi
\ifx \bfpage  \undefined \def \bfpage#1{#1}\fi
\ifx \blpage  \undefined \def \blpage #1{#1}\fi
\ifx \burl  \undefined \def \burl#1{\textsf{#1}}\fi
\ifx \doiurl  \undefined \def \doiurl#1{\url{https://doi.org/#1}}\fi
\ifx \betal  \undefined \def \betal{\textit{et al.}}\fi
\ifx \binstitute  \undefined \def \binstitute#1{#1}\fi
\ifx \binstitutionaled  \undefined \def \binstitutionaled#1{#1}\fi
\ifx \bctitle  \undefined \def \bctitle#1{#1}\fi
\ifx \beditor  \undefined \def \beditor#1{#1}\fi
\ifx \bpublisher  \undefined \def \bpublisher#1{#1}\fi
\ifx \bbtitle  \undefined \def \bbtitle#1{#1}\fi
\ifx \bedition  \undefined \def \bedition#1{#1}\fi
\ifx \bseriesno  \undefined \def \bseriesno#1{#1}\fi
\ifx \blocation  \undefined \def \blocation#1{#1}\fi
\ifx \bsertitle  \undefined \def \bsertitle#1{#1}\fi
\ifx \bsnm \undefined \def \bsnm#1{#1}\fi
\ifx \bsuffix \undefined \def \bsuffix#1{#1}\fi
\ifx \bparticle \undefined \def \bparticle#1{#1}\fi
\ifx \barticle \undefined \def \barticle#1{#1}\fi
\bibcommenthead
\ifx \bconfdate \undefined \def \bconfdate #1{#1}\fi
\ifx \botherref \undefined \def \botherref #1{#1}\fi
\ifx \url \undefined \def \url#1{\textsf{#1}}\fi
\ifx \bchapter \undefined \def \bchapter#1{#1}\fi
\ifx \bbook \undefined \def \bbook#1{#1}\fi
\ifx \bcomment \undefined \def \bcomment#1{#1}\fi
\ifx \oauthor \undefined \def \oauthor#1{#1}\fi
\ifx \citeauthoryear \undefined \def \citeauthoryear#1{#1}\fi
\ifx \endbibitem  \undefined \def \endbibitem {}\fi
\ifx \bconflocation  \undefined \def \bconflocation#1{#1}\fi
\ifx \arxivurl  \undefined \def \arxivurl#1{\textsf{#1}}\fi
\csname PreBibitemsHook\endcsname

\bibitem{van_der_aalst_time_2011}
\begin{botherref}
\oauthor{\bparticle{van~der} \bsnm{Aalst}, \binits{W.M.P.}},
\oauthor{\bsnm{Schonenberg}, \binits{M.H.}},
\oauthor{\bsnm{Song}, \binits{M.}}:
Time prediction based on process mining
\textbf{36}(2),
450--475.
\doiurl{10.1016/j.is.2010.09.001}.
Accessed 2021-01-06
\end{botherref}
\endbibitem

\bibitem{park_action-oriented_2022}
\begin{barticle}
\bauthor{\bsnm{Park}, \binits{G.}},
\bauthor{\bparticle{van~der} \bsnm{Aalst}, \binits{W.M.P.}}:
\batitle{Action-oriented process mining: bridging the gap between insights and
  actions}.
\bjtitle{Progress in Artificial Intelligence}
(\byear{2022}).
\doiurl{10.1007/s13748-022-00281-7}
\end{barticle}
\endbibitem

\bibitem{hitfox_group_comprehensible_2016}
\begin{botherref}
\oauthor{\bsnm{{Hitfox Group}}},
\oauthor{\bsnm{Breuker}, \binits{D.}},
\oauthor{\bsnm{Matzner}, \binits{M.}},
\oauthor{\bsnm{{University of Muenster}}},
\oauthor{\bsnm{Delfmann}, \binits{P.}},
\oauthor{\bsnm{{University of Koblenz-Landau}}},
\oauthor{\bsnm{Becker}, \binits{J.}},
\oauthor{\bsnm{{University of Muenster}}}:
Comprehensible predictive models for business processes
\textbf{40}(4),
1009--1034.
\doiurl{10.25300/MISQ/2016/40.4.10}
\end{botherref}
\endbibitem

\bibitem{marquez-chamorro_predictive_2018}
\begin{botherref}
\oauthor{\bsnm{Marquez-Chamorro}, \binits{A.E.}},
\oauthor{\bsnm{Resinas}, \binits{M.}},
\oauthor{\bsnm{Ruiz-Cortes}, \binits{A.}}:
Predictive monitoring of business processes: A survey
\textbf{11}(6),
962--977.
\doiurl{10.1109/TSC.2017.2772256}.
Accessed 2021-02-14
\end{botherref}
\endbibitem

\bibitem{breiman1996bagging}
\begin{barticle}
\bauthor{\bsnm{Breiman}, \binits{L.}}:
\batitle{Bagging predictors}.
\bjtitle{Machine learning}
\bvolume{24}(\bissue{2}),
\bfpage{123}--\blpage{140}
(\byear{1996})
\end{barticle}
\endbibitem

\bibitem{XGboost}
\begin{bchapter}
\bauthor{\bsnm{Chen}, \binits{T.}},
\bauthor{\bsnm{Guestrin}, \binits{C.}}:
\bctitle{Xgboost: {A} scalable tree boosting system}.
In: \bbtitle{Proceedings of the 22nd {ACM} {SIGKDD} International Conference on
  Knowledge Discovery and Data Mining, San Francisco, CA, USA, August 13-17,
  2016},
pp. \bfpage{785}--\blpage{794}
(\byear{2016}).
\doiurl{10.1145/2939672.2939785}
\end{bchapter}
\endbibitem

\bibitem{senderovich2017intra}
\begin{bchapter}
\bauthor{\bsnm{Senderovich}, \binits{A.}},
\bauthor{\bsnm{Di~Francescomarino}, \binits{C.}},
\bauthor{\bsnm{Ghidini}, \binits{C.}},
\bauthor{\bsnm{Jorbina}, \binits{K.}},
\bauthor{\bsnm{Maggi}, \binits{F.M.}}:
\bctitle{Intra and inter-case features in predictive process monitoring: A tale
  of two dimensions}.
In: \bbtitle{International Conference on Business Process Management},
pp. \bfpage{306}--\blpage{323}
(\byear{2017}).
\bcomment{Springer}
\end{bchapter}
\endbibitem

\bibitem{teinemaa2019outcome}
\begin{barticle}
\bauthor{\bsnm{Teinemaa}, \binits{I.}},
\bauthor{\bsnm{Dumas}, \binits{M.}},
\bauthor{\bsnm{Rosa}, \binits{M.L.}},
\bauthor{\bsnm{Maggi}, \binits{F.M.}}:
\batitle{Outcome-oriented predictive process monitoring: Review and benchmark}.
\bjtitle{ACM Transactions on Knowledge Discovery from Data (TKDD)}
\bvolume{13}(\bissue{2}),
\bfpage{1}--\blpage{57}
(\byear{2019})
\end{barticle}
\endbibitem

\bibitem{DBLP:journals/dss/EvermannRF17}
\begin{barticle}
\bauthor{\bsnm{Evermann}, \binits{J.}},
\bauthor{\bsnm{Rehse}, \binits{J.}},
\bauthor{\bsnm{Fettke}, \binits{P.}}:
\batitle{Predicting process behaviour using deep learning}.
\bjtitle{Decis. Support Syst.}
\bvolume{100},
\bfpage{129}--\blpage{140}
(\byear{2017}).
\doiurl{10.1016/j.dss.2017.04.003}
\end{barticle}
\endbibitem

\bibitem{ZHOU2017350}
\begin{barticle}
\bauthor{\bsnm{Zhou}, \binits{L.}},
\bauthor{\bsnm{Pan}, \binits{S.}},
\bauthor{\bsnm{Wang}, \binits{J.}},
\bauthor{\bsnm{Vasilakos}, \binits{A.V.}}:
\batitle{Machine learning on big data: Opportunities and challenges}.
\bjtitle{Neurocomputing}
\bvolume{237},
\bfpage{350}--\blpage{361}
(\byear{2017}).
\doiurl{10.1016/j.neucom.2017.01.026}
\end{barticle}
\endbibitem

\bibitem{DBLP:conf/spaa/PourghassemiZLC20}
\begin{bchapter}
\bauthor{\bsnm{Pourghassemi}, \binits{B.}},
\bauthor{\bsnm{Zhang}, \binits{C.}},
\bauthor{\bsnm{Lee}, \binits{J.H.}},
\bauthor{\bsnm{Chandramowlishwaran}, \binits{A.}}:
\bctitle{On the limits of parallelizing convolutional neural networks on gpus}.
In: \bbtitle{{SPAA} '20: 32nd {ACM} Symposium on Parallelism in Algorithms and
  Architectures, Virtual Event, USA, July 15-17, 2020},
pp. \bfpage{567}--\blpage{569}
(\byear{2020}).
\doiurl{10.1145/3350755.3400266}
\end{bchapter}
\endbibitem

\bibitem{10.5555/2671164}
\begin{bbook}
\bauthor{\bsnm{Garca}, \binits{S.}},
\bauthor{\bsnm{Luengo}, \binits{J.}},
\bauthor{\bsnm{Herrera}, \binits{F.}}:
\bbtitle{Data Preprocessing in Data Mining},
(\byear{2014}).
\bcomment{Springer Publishing Company, Incorporated}
\end{bbook}
\endbibitem

\bibitem{10.1023/A:1007626913721}
\begin{barticle}
\bauthor{\bsnm{Wilson}, \binits{D.R.}},
\bauthor{\bsnm{Martinez}, \binits{T.R.}}:
\batitle{Reduction techniques for instance-basedlearning algorithms}.
\bjtitle{Mach. Learn.}
\bvolume{38}(\bissue{3}),
\bfpage{257}--\blpage{286}
(\byear{2000}).
\doiurl{10.1023/A:1007626913721}
\end{barticle}
\endbibitem

\bibitem{mnn}
\begin{barticle}
\bauthor{\bsnm{Wilson}, \binits{D.L.}}:
\batitle{Asymptotic properties of nearest neighbor rules using edited data}.
\bjtitle{Systems, Man and Cybernetics, IEEE Transactions on}
\bvolume{2}(\bissue{3}),
\bfpage{408}--\blpage{421}
(\byear{1972}).
\doiurl{10.1109/TSMC.1972.4309137}
\end{barticle}
\endbibitem

\bibitem{de_leoni_general_2016}
\begin{botherref}
\oauthor{\bparticle{de} \bsnm{Leoni}, \binits{M.}},
\oauthor{\bparticle{van~der} \bsnm{Aalst}, \binits{W.M.P.}},
\oauthor{\bsnm{Dees}, \binits{M.}}:
A general process mining framework for correlating, predicting and clustering
  dynamic behavior based on event logs
\textbf{56},
235--257.
\doiurl{10.1016/j.is.2015.07.003}
\end{botherref}
\endbibitem

\bibitem{process-mining}
\begin{bbook}
\bauthor{\bparticle{van~der} \bsnm{Aalst}, \binits{W.M.P.}}:
\bbtitle{Process Mining - Data Science in Action, Second Edition},
(\byear{2016}).
\doiurl{10.1007/978-3-662-49851-4}.
\bcomment{Springer}
\end{bbook}
\endbibitem

\bibitem{LSTM}
\begin{botherref}
\oauthor{\bsnm{Huang}, \binits{Z.}},
\oauthor{\bsnm{Xu}, \binits{W.}},
\oauthor{\bsnm{Yu}, \binits{K.}}:
Bidirectional {LSTM-CRF} models for sequence tagging.
CoRR
\textbf{abs/1508.01991}
(2015)
{\href{https://arxiv.org/abs/1508.01991}{{arXiv:1508.01991}}}
\end{botherref}
\endbibitem

\bibitem{sani2021event}
\begin{bchapter}
\bauthor{\bsnm{Sani}, \binits{M.F.}},
\bauthor{\bsnm{Vazifehdoostirani}, \binits{M.}},
\bauthor{\bsnm{Park}, \binits{G.}},
\bauthor{\bsnm{Pegoraro}, \binits{M.}},
\bauthor{\bparticle{van} \bsnm{Zelst}, \binits{S.J.}},
\bauthor{\bparticle{van~der} \bsnm{Aalst}, \binits{W.M.P.}}:
\bctitle{Event log sampling for predictive monitoring}.
In: \bbtitle{Process Mining Workshops - {ICPM} 2021 International Workshops,
  Eindhoven, The Netherlands, October 31 - November 4, 2021, Revised Selected
  Papers}.
\bsertitle{Lecture Notes in Business Information Processing},
vol. \bseriesno{433},
pp. \bfpage{154}--\blpage{166}
(\byear{2021}).
\doiurl{10.1007/978-3-030-98581-3\_12}
\end{bchapter}
\endbibitem

\bibitem{DBLP:journals/computing/PolatoSBL18}
\begin{barticle}
\bauthor{\bsnm{Polato}, \binits{M.}},
\bauthor{\bsnm{Sperduti}, \binits{A.}},
\bauthor{\bsnm{Burattin}, \binits{A.}},
\bauthor{\bparticle{de} \bsnm{Leoni}, \binits{M.}}:
\batitle{Time and activity sequence prediction of business process instances}.
\bjtitle{Computing}
\bvolume{100}(\bissue{9}),
\bfpage{1005}--\blpage{1031}
(\byear{2018}).
\doiurl{10.1007/s00607-018-0593-x}
\end{barticle}
\endbibitem

\bibitem{DBLP:journals/dss/ParkS20}
\begin{botherref}
\oauthor{\bsnm{Park}, \binits{G.}},
\oauthor{\bsnm{Song}, \binits{M.}}:
Predicting performances in business processes using deep neural networks.
Decis. Support Syst.
\textbf{129}
(2020).
\doiurl{10.1016/j.dss.2019.113191}
\end{botherref}
\endbibitem

\bibitem{DBLP:conf/icsoc/Rogge-SoltiW13}
\begin{bchapter}
\bauthor{\bsnm{Rogge{-}Solti}, \binits{A.}},
\bauthor{\bsnm{Weske}, \binits{M.}}:
\bctitle{Prediction of remaining service execution time using stochastic petri
  nets with arbitrary firing delays}.
In: \beditor{\bsnm{Basu}, \binits{S.}},
\beditor{\bsnm{Pautasso}, \binits{C.}},
\beditor{\bsnm{Zhang}, \binits{L.}},
\beditor{\bsnm{Fu}, \binits{X.}} (eds.)
\bbtitle{Service-Oriented Computing - 11th International Conference, {ICSOC}
  2013, Berlin, Germany, December 2-5, 2013, Proceedings},
vol. \bseriesno{8274},
pp. \bfpage{389}--\blpage{403}
(\byear{2013}).
\doiurl{10.1007/978-3-642-45005-1\_27}
\end{bchapter}
\endbibitem

\bibitem{teinemaa2016predictive}
\begin{bchapter}
\bauthor{\bsnm{Teinemaa}, \binits{I.}},
\bauthor{\bsnm{Dumas}, \binits{M.}},
\bauthor{\bsnm{Maggi}, \binits{F.M.}},
\bauthor{\bsnm{Di~Francescomarino}, \binits{C.}}:
\bctitle{Predictive business process monitoring with structured and
  unstructured data}.
In: \bbtitle{International Conference on Business Process Management},
pp. \bfpage{401}--\blpage{417}
(\byear{2016}).
\bcomment{Springer}
\end{bchapter}
\endbibitem

\bibitem{DBLP:conf/caise/TaxVRD17}
\begin{bchapter}
\bauthor{\bsnm{Tax}, \binits{N.}},
\bauthor{\bsnm{Verenich}, \binits{I.}},
\bauthor{\bsnm{Rosa}, \binits{M.L.}},
\bauthor{\bsnm{Dumas}, \binits{M.}}:
\bctitle{Predictive business process monitoring with {LSTM} neural networks}.
In: \beditor{\bsnm{Dubois}, \binits{E.}},
\beditor{\bsnm{Pohl}, \binits{K.}} (eds.)
\bbtitle{Advanced Information Systems Engineering - 29th International
  Conference, CAiSE 2017, Essen, Germany, June 12-16, 2017, Proceedings},
vol. \bseriesno{10253},
pp. \bfpage{477}--\blpage{492}
(\byear{2017}).
\doiurl{10.1007/978-3-319-59536-8\_30}
\end{bchapter}
\endbibitem

\bibitem{DBLP:conf/ssci/NavarinVPS17}
\begin{bchapter}
\bauthor{\bsnm{Navarin}, \binits{N.}},
\bauthor{\bsnm{Vincenzi}, \binits{B.}},
\bauthor{\bsnm{Polato}, \binits{M.}},
\bauthor{\bsnm{Sperduti}, \binits{A.}}:
\bctitle{{LSTM} networks for data-aware remaining time prediction of business
  process instances}.
In: \bbtitle{2017 {IEEE} Symposium Series on Computational Intelligence, {SSCI}
  2017, Honolulu, HI, USA, November 27 - Dec. 1, 2017},
pp. \bfpage{1}--\blpage{7}
(\byear{2017}).
\doiurl{10.1109/SSCI.2017.8285184}
\end{bchapter}
\endbibitem

\bibitem{DBLP:conf/icpm/NguyenCWSME20}
\begin{bchapter}
\bauthor{\bsnm{Nguyen}, \binits{A.}},
\bauthor{\bsnm{Chatterjee}, \binits{S.}},
\bauthor{\bsnm{Weinzierl}, \binits{S.}},
\bauthor{\bsnm{Schwinn}, \binits{L.}},
\bauthor{\bsnm{Matzner}, \binits{M.}},
\bauthor{\bsnm{Eskofier}, \binits{B.M.}}:
\bctitle{Time matters: Time-aware lstms for predictive business process
  monitoring}.
In: \beditor{\bsnm{Leemans}, \binits{S.J.J.}},
\beditor{\bsnm{Leopold}, \binits{H.}} (eds.)
\bbtitle{Process Mining Workshops - {ICPM} 2020 International Workshops, Padua,
  Italy, October 5-8, 2020, Revised Selected Papers},
vol. \bseriesno{406},
pp. \bfpage{112}--\blpage{123}
(\byear{2020}).
\doiurl{10.1007/978-3-030-72693-5\_9}
\end{bchapter}
\endbibitem

\bibitem{DBLP:conf/bis/PegoraroUGA21}
\begin{bchapter}
\bauthor{\bsnm{Pegoraro}, \binits{M.}},
\bauthor{\bsnm{Uysal}, \binits{M.S.}},
\bauthor{\bsnm{Georgi}, \binits{D.B.}},
\bauthor{\bparticle{van~der} \bsnm{Aalst}, \binits{W.M.P.}}:
\bctitle{Text-aware predictive monitoring of business processes}.
In: \bbtitle{24th International Conference on Business Information Systems,
  {BIS} 2021, Hannover, Germany, June 15-17, 2021},
pp. \bfpage{221}--\blpage{232}
(\byear{2021}).
\doiurl{10.52825/bis.v1i.62}
\end{bchapter}
\endbibitem

\bibitem{DBLP:journals/is/AaRL21}
\begin{barticle}
\bauthor{\bparticle{van~der} \bsnm{Aa}, \binits{H.}},
\bauthor{\bsnm{Rebmann}, \binits{A.}},
\bauthor{\bsnm{Leopold}, \binits{H.}}:
\batitle{Natural language-based detection of semantic execution anomalies in
  event logs}.
\bjtitle{Information Systems}
\bvolume{102},
\bfpage{101824}
(\byear{2021}).
\doiurl{10.1016/j.is.2021.101824}
\end{barticle}
\endbibitem

\bibitem{DBLP:conf/bpmds/0001UHA22}
\begin{bchapter}
\bauthor{\bsnm{Pegoraro}, \binits{M.}},
\bauthor{\bsnm{Uysal}, \binits{M.S.}},
\bauthor{\bsnm{H{\"{u}}lsmann}, \binits{T.}},
\bauthor{\bparticle{van~der} \bsnm{Aalst}, \binits{W.M.P.}}:
\bctitle{Uncertain case identifiers in process mining: {A} user study of the
  event-case correlation problem on click data}.
In: \bbtitle{Enterprise, Business-Process and Information Systems Modeling -
  23rd International Conference, {BPMDS} 2022 and 27th International
  Conference, {EMMSAD} 2022, Held at CAiSE 2022, Leuven, Belgium, June 6-7,
  2022, Proceedings}.
\bsertitle{Lecture Notes in Business Information Processing},
vol. \bseriesno{450},
pp. \bfpage{173}--\blpage{187}
(\byear{2022}).
\doiurl{10.1007/978-3-031-07475-2\_12}
\end{bchapter}
\endbibitem

\bibitem{DBLP:journals/corr/abs-2212-00009}
\begin{botherref}
\oauthor{\bsnm{Pegoraro}, \binits{M.}},
\oauthor{\bsnm{Uysal}, \binits{M.S.}},
\oauthor{\bsnm{H{\"{u}}lsmann}, \binits{T.}},
\oauthor{\bparticle{van~der} \bsnm{Aalst}, \binits{W.M.P.}}:
Resolving uncertain case identifiers in interaction logs: A user study.
CoRR
\textbf{abs/2212.00009}
(2022)
{\href{https://arxiv.org/abs/2212.00009}{{2212.00009}}}.
\doiurl{10.48550/arXiv.2212.00009}
\end{botherref}
\endbibitem

\bibitem{DBLP:conf/bpm/Verenich19}
\begin{bchapter}
\bauthor{\bsnm{Verenich}, \binits{I.}}:
\bctitle{Explainable predictive monitoring of temporal measures of business
  processes}.
In: \bbtitle{Proceedings of the Dissertation Award, Doctoral Consortium, and
  Demonstration Track at {BPM} 2019 Co-located with 17th International
  Conference on Business Process Management, {BPM} 2019, Vienna, Austria,
  September 1-6, 2019}.
\bsertitle{{CEUR} Workshop Proceedings},
vol. \bseriesno{2420},
pp. \bfpage{26}--\blpage{30}
(\byear{2019}).
\burl{http://ceur-ws.org/Vol-2420/paperDA6.pdf}
\end{bchapter}
\endbibitem

\bibitem{DBLP:conf/ecis/StierleBWZM021}
\begin{bchapter}
\bauthor{\bsnm{Stierle}, \binits{M.}},
\bauthor{\bsnm{Brunk}, \binits{J.}},
\bauthor{\bsnm{Weinzierl}, \binits{S.}},
\bauthor{\bsnm{Zilker}, \binits{S.}},
\bauthor{\bsnm{Matzner}, \binits{M.}},
\bauthor{\bsnm{Becker}, \binits{J.}}:
\bctitle{Bringing light into the darkness - {A} systematic literature review on
  explainable predictive business process monitoring techniques}.
In: \bbtitle{28th European Conference on Information Systems - Liberty,
  Equality, and Fraternity in a Digitizing World , {ECIS} 2020, Marrakech,
  Morocco, June 15-17, 2020}
(\byear{2021}).
\burl{https://aisel.aisnet.org/ecis2021\_rip/8}
\end{bchapter}
\endbibitem

\bibitem{DBLP:conf/bpm/SindhgattaMOB20}
\begin{bchapter}
\bauthor{\bsnm{Sindhgatta}, \binits{R.}},
\bauthor{\bsnm{Moreira}, \binits{C.}},
\bauthor{\bsnm{Ouyang}, \binits{C.}},
\bauthor{\bsnm{Barros}, \binits{A.}}:
\bctitle{Exploring interpretable predictive models for business processes}.
In: \bbtitle{Business Process Management - 18th International Conference, {BPM}
  2020, Seville, Spain, September 13-18, 2020, Proceedings}.
\bsertitle{Lecture Notes in Computer Science},
vol. \bseriesno{12168},
pp. \bfpage{257}--\blpage{272}
(\byear{2020}).
\doiurl{10.1007/978-3-030-58666-9\_15}
\end{bchapter}
\endbibitem

\bibitem{DBLP:conf/icpm/GalantiCLCN20}
\begin{bchapter}
\bauthor{\bsnm{Galanti}, \binits{R.}},
\bauthor{\bsnm{Coma{-}Puig}, \binits{B.}},
\bauthor{\bparticle{de} \bsnm{Leoni}, \binits{M.}},
\bauthor{\bsnm{Carmona}, \binits{J.}},
\bauthor{\bsnm{Navarin}, \binits{N.}}:
\bctitle{Explainable predictive process monitoring}.
In: \bbtitle{2nd International Conference on Process Mining, {ICPM} 2020,
  Padua, Italy, October 4-9, 2020},
pp. \bfpage{1}--\blpage{8}
(\byear{2020}).
\doiurl{10.1109/ICPM49681.2020.00012}
\end{bchapter}
\endbibitem

\bibitem{DBLP:journals/corr/abs-2210-16786}
\begin{botherref}
\oauthor{\bsnm{Park}, \binits{G.}},
\oauthor{\bsnm{K{\"{u}}sters}, \binits{A.}},
\oauthor{\bsnm{Tews}, \binits{M.}},
\oauthor{\bsnm{Pitsch}, \binits{C.}},
\oauthor{\bsnm{Schneider}, \binits{J.}},
\oauthor{\bparticle{van~der} \bsnm{Aalst}, \binits{W.M.P.}}:
Explainable predictive decision mining for operational support.
CoRR
\textbf{abs/2210.16786}
(2022)
{\href{https://arxiv.org/abs/2210.16786}{{2210.16786}}}.
\doiurl{10.48550/arXiv.2210.16786}
\end{botherref}
\endbibitem

\bibitem{DBLP:conf/bpm/PauwelsC21}
\begin{bchapter}
\bauthor{\bsnm{Pauwels}, \binits{S.}},
\bauthor{\bsnm{Calders}, \binits{T.}}:
\bctitle{Incremental predictive process monitoring: The next activity case}.
In: \bbtitle{Business Process Management - 19th International Conference, {BPM}
  2021, Rome, Italy, September 06-10, 2021, Proceedings}.
\bsertitle{Lecture Notes in Computer Science},
vol. \bseriesno{12875},
pp. \bfpage{123}--\blpage{140}
(\byear{2021}).
\doiurl{10.1007/978-3-030-85469-0\_10}
\end{bchapter}
\endbibitem

\bibitem{wang2020dataset}
\begin{botherref}
\oauthor{\bsnm{Wang}, \binits{T.}},
\oauthor{\bsnm{Zhu}, \binits{J.-Y.}},
\oauthor{\bsnm{Torralba}, \binits{A.}},
\oauthor{\bsnm{Efros}, \binits{A.A.}}:
Dataset distillation.
arXiv preprint arXiv:1811.10959
(2020)
{\href{https://arxiv.org/abs/1811.10959}{{arXiv:1811.10959}}}
{[cs.LG]}
\end{botherref}
\endbibitem

\bibitem{sani_2021_sampling}
\begin{barticle}
\bauthor{\bsnm{Fani~Sani}, \binits{M.}},
\bauthor{\bparticle{van} \bsnm{Zelst}, \binits{S.J.}},
\bauthor{\bparticle{van~der} \bsnm{Aalst}, \binits{W.M.P.}}:
\batitle{The impact of biased sampling of event logs on the performance of
  process discovery}.
\bjtitle{Computing}
\bvolume{103}(\bissue{6}),
\bfpage{1085}--\blpage{1104}
(\byear{2021}).
\doiurl{10.1007/s00607-021-00910-4}
\end{barticle}
\endbibitem

\bibitem{sani_2020_editDistance}
\begin{bchapter}
\bauthor{\bsnm{Fani~Sani}, \binits{M.}},
\bauthor{\bparticle{van} \bsnm{Zelst}, \binits{S.J.}},
\bauthor{\bparticle{van~der} \bsnm{Aalst}, \binits{W.M.P.}}:
\bctitle{Conformance checking approximation using subset selection and edit
  distance}.
In: \bbtitle{Advanced Information Systems Engineering - 32nd International
  Conference, CAiSE 2020, Grenoble, France, June 8-12, 2020, Proceedings},
vol. \bseriesno{12127},
pp. \bfpage{234}--\blpage{251}
(\byear{2020}).
\doiurl{10.1007/978-3-030-49435-3\_15}
\end{bchapter}
\endbibitem

\bibitem{qafari2020feature}
\begin{bchapter}
\bauthor{\bsnm{Qafari}, \binits{M.S.}},
\bauthor{\bparticle{van~der} \bsnm{Aalst}, \binits{W.M.P.}}:
\bctitle{Root cause analysis in process mining using structural equation
  models}.
In: \bbtitle{Business Process Management Workshops - {BPM} 2020 International
  Workshops, Seville, Spain, September 13-18, 2020, Revised Selected Papers},
vol. \bseriesno{397},
pp. \bfpage{155}--\blpage{167}
(\byear{2020}).
\doiurl{10.1007/978-3-030-66498-5\_12}
\end{bchapter}
\endbibitem

\bibitem{Prom}
\begin{bchapter}
\bauthor{\bsnm{Verbeek}, \binits{E.}},
\bauthor{\bsnm{Buijs}, \binits{J.C.A.M.}},
\bauthor{\bparticle{van} \bsnm{Dongen}, \binits{B.F.}},
\bauthor{\bparticle{van~der} \bsnm{Aalst}, \binits{W.M.P.}}:
\bctitle{Prom 6: The process mining toolkit}.
In: \bbtitle{Proceedings of the Business Process Management 2010 Demonstration
  Track, Hoboken, NJ, USA, September 14-16, 2010},
vol. \bseriesno{615}
(\byear{2010}).
\burl{http://ceur-ws.org/Vol-615/paper13.pdf}
\end{bchapter}
\endbibitem

\bibitem{DBLP:conf/icpm/ParkS19}
\begin{bchapter}
\bauthor{\bsnm{Park}, \binits{G.}},
\bauthor{\bsnm{Song}, \binits{M.}}:
\bctitle{Prediction-based resource allocation using {LSTM} and minimum cost and
  maximum flow algorithm}.
In: \bbtitle{International Conference on Process Mining, {ICPM} 2019, Aachen,
  Germany, June 24-26, 2019},
pp. \bfpage{121}--\blpage{128}
(\byear{2019}).
\doiurl{10.1109/ICPM.2019.00027}
\end{bchapter}
\endbibitem

\bibitem{verenich2019survey}
\begin{barticle}
\bauthor{\bsnm{Verenich}, \binits{I.}},
\bauthor{\bsnm{Dumas}, \binits{M.}},
\bauthor{\bsnm{Rosa}, \binits{M.L.}},
\bauthor{\bsnm{Maggi}, \binits{F.M.}},
\bauthor{\bsnm{Teinemaa}, \binits{I.}}:
\batitle{Survey and cross-benchmark comparison of remaining time prediction
  methods in business process monitoring}.
\bjtitle{ACM Transactions on Intelligent Systems and Technology (TIST)}
\bvolume{10}(\bissue{4}),
\bfpage{1}--\blpage{34}
(\byear{2019})
\end{barticle}
\endbibitem

\bibitem{DBLP:journals/pr/LuqueCMH19}
\begin{barticle}
\bauthor{\bsnm{Luque}, \binits{A.}},
\bauthor{\bsnm{Carrasco}, \binits{A.}},
\bauthor{\bsnm{Mart{\'{\i}}n}, \binits{A.}},
\bauthor{\bparticle{de~las} \bsnm{Heras}, \binits{A.}}:
\batitle{The impact of class imbalance in classification performance metrics
  based on the binary confusion matrix}.
\bjtitle{Pattern Recognit.}
\bvolume{91},
\bfpage{216}--\blpage{231}
(\byear{2019}).
\doiurl{10.1016/j.patcog.2019.02.023}
\end{barticle}
\endbibitem

\bibitem{Irene_outcome_2019}
\begin{barticle}
\bauthor{\bsnm{Teinemaa}, \binits{I.}},
\bauthor{\bsnm{Dumas}, \binits{M.}},
\bauthor{\bsnm{Rosa}, \binits{M.L.}},
\bauthor{\bsnm{Maggi}, \binits{F.M.}}:
\batitle{Outcome-oriented predictive process monitoring: Review and benchmark}.
\bjtitle{{ACM} Trans. Knowl. Discov. Data}
\bvolume{13}(\bissue{2}),
\bfpage{17}--\blpage{11757}
(\byear{2019}).
\doiurl{10.1145/3301300}
\end{barticle}
\endbibitem

\bibitem{Hyperparameter}
\begin{bchapter}
\bauthor{\bsnm{Bergstra}, \binits{J.}},
\bauthor{\bsnm{Bardenet}, \binits{R.}},
\bauthor{\bsnm{Bengio}, \binits{Y.}},
\bauthor{\bsnm{K{\'{e}}gl}, \binits{B.}}:
\bctitle{Algorithms for hyper-parameter optimization}.
In: \bbtitle{Advances in Neural Information Processing Systems 24: 25th Annual
  Conference on Neural Information Processing Systems 2011. Proceedings of a
  Meeting Held 12-14 December 2011, Granada, Spain},
pp. \bfpage{2546}--\blpage{2554}
(\byear{2011})
\end{bchapter}
\endbibitem

\bibitem{josep_2012_drift}
\begin{bchapter}
\bauthor{\bsnm{Carmona}, \binits{J.}},
\bauthor{\bsnm{Gavald{\`{a}}}, \binits{R.}}:
\bctitle{Online techniques for dealing with concept drift in process mining}.
In: \bbtitle{Advances in Intelligent Data Analysis {XI} - 11th International
  Symposium, {IDA} 2012, Helsinki, Finland, October 25-27, 2012. Proceedings},
vol. \bseriesno{7619},
pp. \bfpage{90}--\blpage{102}
(\byear{2012}).
\doiurl{10.1007/978-3-642-34156-4\_10}
\end{bchapter}
\endbibitem

\bibitem{bias}
\begin{barticle}
\bauthor{\bparticle{van~der} \bsnm{Putten}, \binits{P.}},
\bauthor{\bparticle{van} \bsnm{Someren}, \binits{M.}}:
\batitle{A bias-variance analysis of a real world learning problem: The coil
  challenge 2000}.
\bjtitle{Mach. Learn.}
\bvolume{57}(\bissue{1-2}),
\bfpage{177}--\blpage{195}
(\byear{2004}).
\doiurl{10.1023/B:MACH.0000035476.95130.99}
\end{barticle}
\endbibitem

\end{thebibliography}
\end{document}